\documentclass[final]{cvpr}

\usepackage{times}
\usepackage{epsfig}
\usepackage{graphicx}
\usepackage{amsmath}
\usepackage{amssymb}

\usepackage{url}
\usepackage{multirow}
\usepackage{makecell}
\usepackage{placeins}
\usepackage{xcolor}
\usepackage{supertabular,booktabs}
\usepackage{tabularx}
\usepackage{balance}
\usepackage{longtable}
\usepackage{enumitem}

\usepackage[acronym,nonumberlist,nogroupskip,nopostdot]{glossaries}

\usepackage[pagebackref=true,breaklinks=true,colorlinks,bookmarks=false]{hyperref}

\newacronym{objectnav}{ObjectNav}{Object-Goal Navigation}
\newacronym{ai}{AI}{Artificial Intelligence}
\newacronym{eai}{E-AI}{Embodied Artificial Intelligence}
\newacronym{rl}{RL}{Reinforcement Learning}
\newacronym{deeprl}{Deep-RL}{Deep Reinforcement Learning}
\newacronym{lstm}{LSTM}{Long Short-Term Memory}
\newacronym{rnn}{RNN}{Recurrent Neural Network}
\newacronym{ddppo}{DD-PPO}{Decentralized Distributed Proximal Policy Optimization}
\newacronym{ppo}{PPO}{Proximal Policy Optimization}

\def\cvprPaperID{8} 
\def\confYear{CVPR 2021}

\begin{document}

\title{BEyond observation: an approach for ObjectNav}


\author{Daniel V. Ruiz and Eduardo Todt\\
Department of Informatics, Federal University of Paraná\\
Brazil, Paraná, Curitiba\\
{\tt\small dvruiz@inf.ufpr.br and \tt\small todt@inf.ufpr.br}
}

\maketitle

\begin{abstract}
    With the rise of automation, unmanned vehicles became a hot topic both as commercial products and as a scientific research topic. It composes a multi-disciplinary field of robotics that encompasses embedded systems, control theory, path planning, Simultaneous Localization and Mapping (SLAM), scene reconstruction, and pattern recognition. In this work, we present our exploratory research of how sensor data fusion and state-of-the-art machine learning algorithms can perform the Embodied Artificial Intelligence (E-AI) task called Visual Semantic Navigation. This task, a.k.a Object-Goal  Navigation  (ObjectNav) consists of autonomous navigation using egocentric visual observations to reach an object belonging to the target semantic class without prior knowledge of the environment. Our method reached fourth place on the Habitat Challenge 2021 ObjectNav on the Minival phase and the Test-Standard Phase.
\end{abstract}


\section{Proposed Work}
The first step is to perform semantic segmentation on the egocentric RGB observation at step $t$. In our work, we use the Yolact++ architecture for this task~\cite{bolya2020yolactplus}. We used transfer learning to fine-tune the neural network to the 21 classes on the Habitat challenge 2021 \gls{objectnav}. Those 21 classes are a subset of the original 40 classes of the Matterport3D dataset~\cite{matterport3d}.

The confidence threshold for the pixel-wise prediction chosen was empirically set to 0.55. The source code for our method is publicly available at~\footnote{\url{https://github.com/VRI-UFPR/BeyondSight/tree/beyond_habitat_challenge_2021}}. The semantic prediction is then projected to a 2D top-view map using the depth image at step $t$ and the GPS and Compass sensors measurements available at step $t$. The former provides the current agent's position in an episodic-based coordinate system, where the agent always starts at the origin regardless of its spawning world coordinates. The latter provide the agent's yaw orientation in radians in an episodic coordinate system, where the starting orientation is always 0.

We chose a squared map representation of 512x512 cells, where each squared cell has a 0.05 meter resolution on world scale.  The map has 25 channels, where channel 0 represents obstacles observed by the depth camera without a semantic class. Channels 1 to 21 are dedicated to representing the Habitat challenge 2021 \gls{objectnav} classes. Channel 22 is a max-pooling of all previous channels to provide an occupation map regardless of class. Channel 23 contains all the agent's locations starting from step 0. Channel 24 contains only the agent's current location. This map is centered cropped around the agent's current location to 25x256x256, and padded with zeros if necessary, and then fed along with the current episode target class ID and agent's current orientation to our policy neural network named BEyond.

\begin{figure*}[!h]
    \centering
    \includegraphics[width=0.8\linewidth]{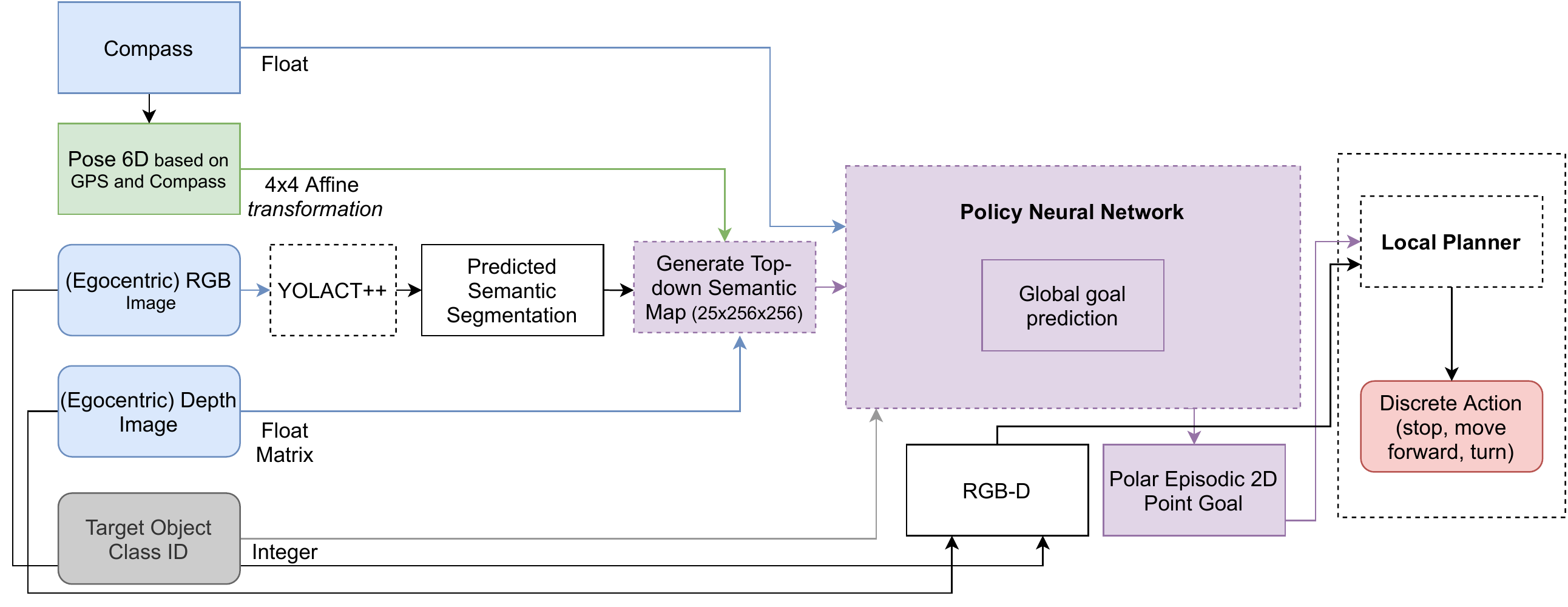}
    \caption[BEyond flowchart]{BEyond flowchart. In green is the resulting pose using GPS and Compass data provided by the simulation. In blue are the Compass value, RGB, and depth images observed at step $t$. In gray the target class ID provided by the episode's settings. In purple are the module that creates the top-down semantic map, our policy neural network that performs the global goal prediction and the subsequent transformation to episodic polar coordinates. In red is the output containing the predicted action.}
    \label{fig:beyond}
\end{figure*}

BEyond follows an Actor-Critic architecture and was trained using \gls{deeprl} on the Habitat Simulator~\cite{savva2019habitat} for around 1 million steps using the \gls{ppo} algorithm~\cite{schulman2017ppo}. The proposed architecture was inspired by ~\cite{beyond2021}. We processed it in a single GPU, an NVIDIA Titan XP. Our policy network operates akin to a global planner instead of a local planner, meaning that it does not output discrete actions to be used on the subsequent step of the simulation. Instead, it provides as output two continuous coordinates. These coordinates are samples from a Gaussian distribution described by the actor head layer and auxiliary learned parameters. The predictions are normalized to the range $[0,1]$ using a sigmoid function. The flowchart of our method is presented in Figure~\ref{fig:beyond}. We tested two approaches of representation for these coordinates: Cartesian and Polar.

\subsection{Cartesian}
The prediction is multiplied by the map's size, then rounded to an integer, representing the agent's current global goal in cell coordinates. Next, the local goal is computed using the $A^{*}$ algorithm to the closest non-obstruct step toward the goal using channel 22. This local goal is transformed to episodic polar coordinates and fed into a different pre-trained actor-critic network that acts as a  local planner. This network receives an RGB-D image and the local goal as input and outputs a discrete action within the possible ones (STOP, MOVE\_FORWARD, TURN\_LEFT, TURN\_RIGHT). This decoupling into two modules, global planner and local planner, allows each component to be easily replaced by another method.

\subsection{Polar}
In this representation, the first coordinate represents the distance $\rho$ to the target, and the second coordinate the orientation $\phi$. The latter is transformed from the [0,1] range to the $[-\pi,+\pi]$ range. The global goal is then fed to the same local planner as the Cartesian one.

The local planner was trained using the \gls{ddppo} method~\cite{wijmans2020ddppo}, and it is the same for both cases.

\section{Reward Function}
In \gls{deeprl} the reward function is essential for learning. We use $-10^{-4}$ as the default value for a step, and $10^{-3}$ if the policy action induced the local planner to take the same discrete action as a greedy oracle with perfect knowledge about the scene. Previous distance to goal minus the current one is another component, and finally, if the episode ended in success, an additional 2.5 (10 times the step size of forward movements) value is added to the reward. This avoids sparse rewards and introduces an imitation learning coefficient in the reward.


\section{Habitat Challenge 2021}
We submitted the variant of our method that uses the Cartesian representation to the Habitat Challenge 2021 in the ObjectNav category Minival phase, see Table~\ref{tab:objectnavChallenge2021MiniVal}. Additionally, the variant that uses the Polar representation was submitted to the Test-Standard phase, see Table~\ref{tab:objectnavChallenge2021Test}.

\begin{table}[!h]
\resizebox{\linewidth}{!}{
\begin{tabular}{cccccc}
\toprule
\multicolumn{1}{c}{\textbf{Rank}} & \multicolumn{1}{c}{\textbf{Team}} & \multicolumn{1}{c}{\textbf{SPL $\uparrow$}} & \multicolumn{1}{c}{\textbf{SoftSPL $\uparrow$}} & \multicolumn{1}{c}{\textbf{Distance to goal $\downarrow$}} & \multicolumn{1}{c}{\textbf{Success rate$\uparrow$}} \\
\midrule
1 & TreasureHunt & 0.15 & 0.25 & 3.41 & 0.27  \\
2 & AIstar (RL) & 0.09 & 0.16 & 3.32 & 0.23  \\
3 & Clueless-Wanderers (Peter) & 0.03 & 0.10 & 4.41 & 0.13  \\
4 & \textbf{BEyond-VRI-UFPR} & \textbf{0.00} & \textbf{0.14} & \textbf{5.71} & \textbf{0.00}  \\
5 & See through pixels (init) & 0.00 & 0.00 & 6.39 & 0.00  \\
6 & Black Swan & 0.00 & 0.01 & 6.38 & 0.00  \\
 \bottomrule
\end{tabular}}
\caption[Habitat Challenge 2021: ObjectNav Leaderboard]{Habitat Challenge 2021: ObjectNav Leaderboard Minival Phase, ranked by SPL ~\cite{habitat2021LeaderboardMini}.}
\label{tab:objectnavChallenge2021MiniVal}
\end{table}

\begin{table}[!h]
\resizebox{\linewidth}{!}{
\begin{tabular}{cccccc}
\toprule
\multicolumn{1}{c}{\textbf{Rank}} & \multicolumn{1}{c}{\textbf{Team}} & \multicolumn{1}{c}{\textbf{SPL $\uparrow$}} & \multicolumn{1}{c}{\textbf{SoftSPL $\uparrow$}} & \multicolumn{1}{c}{\textbf{Distance to goal $\downarrow$}} & \multicolumn{1}{c}{\textbf{Success rate$\uparrow$}} \\
\midrule
1 & TreasureHunt & 0.09 & 0.17 & 9.23 & 0.21  \\
2 & AIstar (RL) & 0.03 & 0.11 & 9.41 & 0.10  \\
3 & Clueless-Wanderers (Peter) & 0.02 & 0.10 & 9.07 & 0.07  \\
4 & \textbf{BEyond-VRI-UFPR} & \textbf{0.00} & \textbf{0.08} & \textbf{10.18} & \textbf{0.00}  \\
5 & Habitat Team (RGBD+DD-PPO) & 0.00 & 0.01 & 10.33 & 0.00  \\
 \bottomrule
\end{tabular}}
\caption[Habitat Challenge 2021: ObjectNav Leaderboard]{Habitat Challenge 2021: ObjectNav Leaderboard Test-Standard Phase, ranked by SPL~\cite{habitat2021LeaderboardTest}. Habitat Team (RGBD+DD-PPO) is the baseline. }
\label{tab:objectnavChallenge2021Test}
\end{table}

The version with the Polar representation had similar results on the Minival phase obtaining (`spl': 0.0, `softspl': 0.12, `distance to goal': 5.75, `success rate': 0.0). 

\section{Conclusion}

We achieved fourth place on the Minival phase and Test-Standard phase,  marking our method as a promising approach for dealing with \gls{objectnav}, but with a large room for future improvements. The main aspects of our method are: the use of semantic segmentation, the projection to a top-down semantic 2D grid, the decoupled global and local planner, and the modified reward function combining reinforcement and imitation learning.


{\small
\bibliographystyle{ieee_fullname}
\bibliography{egbib}
}

\end{document}